\documentclass[runningheads]{llncs}

\usepackage{eccv}

\usepackage{eccvabbrv}

\usepackage{graphicx}
\usepackage{booktabs}
\usepackage{multirow}
\usepackage{amsmath}
\usepackage{amssymb}
\usepackage{mathtools}
\usepackage{tabularx}
\usepackage{algorithm}
\usepackage{makecell}
\usepackage{xcolor,colortbl}
\usepackage{algpseudocode}
\usepackage[accsupp]{axessibility}  

\DeclareMathOperator*{\TopK}{TopK}
\DeclareMathOperator*{\avg}{avg}

\definecolor{Periwinkle}{rgb}{0.56,0.56,0.86}

\usepackage{hyperref}

\usepackage{orcidlink}

\begin{document}

\title{Curing Semantic Drift: A Dynamic Approach to Grounding Generation in LVLMs}

\titlerunning{Curing Semantic Drift}

\author{Jiahe Chen\inst{1} \and
Jiaying He\inst{1} \and
Qiyuan Chen\inst{1} \and
Qian Shao\inst{1} \and
Jiahe Ying\inst{4} \and
Hongxia Xu\inst{1} \and
Jintai Chen\inst{3} \and
Jianwei Zheng\inst{2} \and
Jian Wu\inst{1}
}

\authorrunning{J.~Chen et al.}

\institute{Zhejiang University \and
Zhejiang University of Technology \and
Hong Kong University of Science and Technology \and
Fudan University}

\maketitle

\begin{abstract}
Large Vision-Language Models (LVLMs) face a tug-of-war between powerful linguistic priors and visual evidence, often leading to \emph{semantic drift}: a progressive detachment from the input image that can abruptly emerge at specific decoding steps.
Through a token-level diagnosis, we show that hallucination is frequently triggered not by the absence of grounded candidates, but by a failure of selection---the model chooses a linguistically convenient yet visually unfaithful token even when better grounded alternatives exist.
Motivated by this insight, we propose \textbf{D}ynamic \textbf{L}ogits \textbf{C}alibration (DLC), a training-free decoding framework that introduces a lightweight visual referee to intervene exactly when drift happens.
At each step, DLC performs a dual-aspect grounding check on top-$k$ candidates: (1) it assesses the intrinsic visual relevance of a candidate token and (2) its contextual visual coherence.
These signals are evaluated against an adaptive historical baseline to compute a relative visual advantage, which is then used to dynamically calibrate logits and favor grounded tokens.
Extensive experiments on CHAIR, POPE, SHR, GPT-4o evaluation, and MME demonstrate that DLC consistently reduces hallucinations across multiple LVLMs while preserving response quality.
Further analyses validate robustness to different vision backbones and demonstrate a favorable trade-off between output quality and computational cost as the candidate pool size varies.
\end{abstract}    
\section{Introduction}
\label{sec:intro}


\begin{figure}
    \centering
    \includegraphics[width=0.95\linewidth]{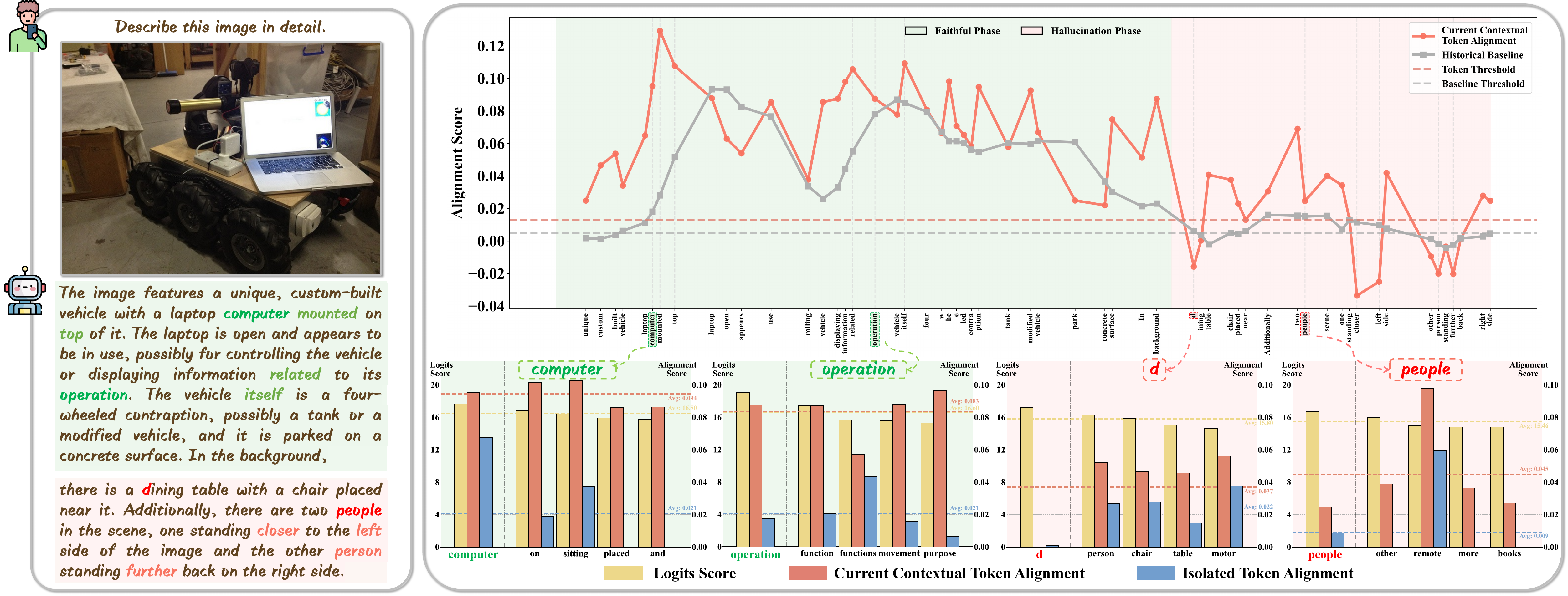}
\caption{\textbf{Semantic drift manifests as a detectable collapse of visual alignment and a failure of token selection.}
The central plot tracks the Current Contextual Token Alignment (CCTA) against an adaptive historical baseline $\bar{B}_t$, contrasting visually faithful (\textbf{\textcolor[HTML]{6CAB42}{green}}) and hallucination (\textbf{\textcolor[HTML]{F87561}{red}}) phases.
The accompanying token-level snapshots compare top-candidate preferences with their visual alignment, revealing that hallucination is often triggered when a linguistically plausible but visually unfaithful token is selected over grounded alternatives.}
\label{fig:motivation}
\vspace{-0.4cm}
\end{figure}

In recent years, LVLMs \cite{shikra,InstructBLIP,llava-next,Blip-2,llava,Qwen2-VL,mplug-owl2,minigpt4} have revolutionized multimodal understanding \cite{Qwen-VL,survey,lisa,liu2024improved} by integrating powerful visual encoders \cite{clip,siglip,fgclip} with Large Language Models (LLMs) \cite{qwen,vicuna,llama,llama2}. However, the impressive autoregressive generation capabilities of these models conceal an inherent conflict: a fundamental tension between following the strong statistical patterns learned from vast text corpora (linguistic priors) and remaining faithful to the visual evidence in the input image \cite{survey,perturbollava,opera,vcd,vasparse}. This conflict gives rise to a phenomenon we term semantic drift: as decoding proceeds—especially for long-form generation—the model’s outputs progressively deviate from the visual context and become increasingly dominated by linguistic priors. This drift is a major driver of hallucination, the generation of plausible but factually ungrounded text, which remains a critical bottleneck for deploying LVLMs in high-stakes domains like medical diagnosis \cite{meddetecting,llavamed} and autonomous systems \cite{driving,reason2drive}.

To cure this drift, we dissect its micro-dynamics and argue that hallucination is not a static error rate but a trajectory failure during decoding. As previewed in Figure \ref{fig:motivation}, a token-level visualization of the Current Contextual Token Alignment (CCTA) reveals two key observations. First, \emph{drift has a moment:} the alignment remains stable when generation is grounded, but collapses sharply at the onset of hallucination, suggesting a detectable transition rather than uniform noise. Second, \emph{drift is often a selection problem:} around these critical steps, visually grounded tokens can still appear among top candidates, yet the model selects a linguistically convenient but visually unfaithful token. This indicates that the model is not necessarily blind, but can be overruled by linguistic priors during token choice. (A detailed analysis is provided in the Section \ref{sec:motivation}.)

This diagnosis motivates a simple principle: if hallucination stems from choosing the wrong token at the right time, then we do not need to retrain the LVLM or perform multi-pass contrastive decoding; instead, we need a mechanism that injects direct visual grounding signals into the token selection process in real time. We propose \textbf{D}ynamic \textbf{L}ogits \textbf{C}alibration (DLC), a lightweight, training-free decoding framework that acts as a visual referee. At each step, DLC evaluates candidate tokens from two complementary aspects—Intrinsic Relevance (is the concept visually present?) and Contextual Coherence (is it consistent with the current visual narrative?)—and dynamically calibrates logits to prevent linguistic priors from hijacking token choice, without disrupting the generation flow.

Our main contributions are threefold:
\begin{itemize}
    \item We provide the first systematic analysis of semantic drift as a progressive, token-level phenomenon, dissecting the mechanism of linguistic prior hijacking and establishing this dynamic failure as the core challenge for real-time, adaptive intervention.
    \item We propose DLC, a novel training-free decoding framework that introduces a dual-aspect visual grounding mechanism (intrinsic relevance and contextual coherence) to effectively cure semantic drift at each token step.
    \item We demonstrate through extensive experiments that DLC sets a new state-of-the-art in hallucination mitigation across various LVLMs and benchmarks, offering a practical path toward trustworthy LVLM decoding.
\end{itemize}

\section{Related Work}

\subsection{The Challenge of Hallucination in LVLMs}
Hallucination in LVLMs occurs when generated text contradicts the visual input, arising from a tension between the model's internal knowledge and external visual evidence. While users expect responses to be strictly conditioned on the image, LVLMs often default to powerful linguistic priors acquired from their training corpus. This tendency causes the generative process to favor statistically likely but visually ungrounded phrases, leading to a progressive semantic drift away from the visual reality as the text sequence unfolds. Our work confronts this core challenge by proposing a mechanism to dynamically re-anchor the generation process to visual evidence at inference time (see Appendix for additional background on LVLMs).



\subsection{A Taxonomy of Mitigation Strategies}
Prior approaches to mitigating hallucination in LVLMs can be broadly grouped by \emph{when} and \emph{how} they intervene in the generation pipeline.
\textbf{1) Data-driven optimization} \cite{perturbollava,hacl,rit,gfaif,rlhf_v,rlaif_v,lessismore,hadpo} improves faithfulness by modifying model behavior via instruction tuning or RLHF. While effective, these methods typically require substantial curated data and careful alignment to avoid overfitting or regressions across tasks.
\textbf{2) Post-hoc correction} \cite{Pelican,LogicCheckGPT,woodpecker,amoh} detects and rectifies errors via auxiliary verifiers, tool use, or external knowledge. Although this improves reliability, it functions reactively—correcting errors only after they occur—which increases system complexity and inference latency.
\textbf{3) Decoding-time refinement} offers a training-free alternative that adjusts generation on the fly. This line can be further categorized by the source of guidance:

\begin{itemize}
\item \textbf{Internal self-guided decoding.}
Approaches such as VCD \cite{vcd}, SID \cite{sid}, and OPERA \cite{opera} exploit signals from the LVLM itself to discourage hallucination. While self-contained, they often require additional LVLM evaluations or specialized decoding procedures, and typically rely on heuristic signals rather than explicit semantic verification.

\item \textbf{External-guided decoding.}
Methods like HALC \cite{halc}, MARINE \cite{marine}, DeGF \cite{degf}, and CGD \cite{gcd} use external guidance to steer generation. Such guidance provides explicit semantic signals but often introduces extra modules or search overhead. Furthermore, many of these methods operate at a coarse granularity, lacking the agility to intervene precisely at the token level.
\end{itemize}

Despite their diversity, many existing methods rely on static penalties or incur prohibitive computation to address what is fundamentally a dynamic, token-level failure mode.
DLC fills this gap by performing direct, token-level calibration with a lightweight visual referee: it injects grounded signals (intrinsic relevance and contextual coherence) into token selection exactly as semantic drift emerges. Crucially, DLC achieves this without the additional forward passes through the LVLM backbone required by multi-pass contrastive decoding baselines, ensuring both efficiency and precision.

\section{Preliminary and Motivation}


\subsection{Generative Process of LVLMs}
An LVLM, parameterized by $\theta$, processes jointly visual images $v$ and textual queries $x$. Typically, a pre-trained network encodes $v$ into visual features. An alignment module (\textit{e.g.}, Q-Former \cite{Blip-2} or linear projection \cite{llava}) then maps these features into the textual semantic space. These aligned visual representations are concatenated with the embedded text query $x$ to form a joint input representation. Conditioned on this unified input, the LVLM auto-regressively predicts the next token $y_t$:
    \begin{equation}
        y_{t}\sim p_{\theta}\left(y_{t}\mid v, x, y_{<t}\right) \propto \exp \left
        ({logit}_{\theta}\left(y_{t}\mid v, x, y_{<t}\right)\right)
    \end{equation}
where $y_{<t}$ are previous tokens and $logit_{\theta}$ represents the model's raw, unnormalized belief scores for the next token.

Decoding strategies then select the next token, which is appended to the input for subsequent iterations. Our proposed DLC enhances this generative process as a training-free decoding strategy, integrating seamlessly by modulating these logits before the final token selection.

\subsection{CLIP Model and CLIP Score}

The Contrastive Language-Image Pre-training (CLIP) model \cite{clip} learns a shared, high-dimensional semantic space for both images and text.
Formally, for a given image $v$ and text sequence $s$, their semantic similarity is computed using cosine similarity, denoted as $\mathrm{CLIP}(v, s)$:
\begin{equation}
\mathrm{CLIP}(v, s) = \frac{E_{I}(v) \cdot E_{T}(s)}{\lVert E_{I}(v) \rVert  \lVert E_{T}(s) \rVert}
\label{eq:clip_score}
\end{equation}
where $E_{I}(v)$ and $E_{T}(s)$ are embeddings from CLIP's pre-trained image and text encoders. A higher $\mathrm{CLIP}(v, s)$ value signifies greater semantic relevance. In our framework, we leverage this score as a quantitative measure of the alignment between a textual hypothesis and the visual evidence.

\subsection{Motivation}\label{sec:motivation}

The progressive emergence of hallucination is a key symptom of semantic drift. To understand its cause, we must move beyond observing the pattern and instead dissect the micro-dynamics of the generation process at the token level. This imperative motivated our fine-grained visual analysis.

\textbf{The Investigation: Tracking Visual Fidelity.} 
To dissect this drift at a granular level, we employ the CLIP score as a proxy to measure the semantic alignment between the input image $v$ and the evolving text sequence $y_t$. Our analysis, visualized in Figure~\ref{fig:motivation}, is twofold:

\textbf{(1) A Macro-level Trajectory:} The line plot tracks the overall visual consistency of the generation over time, allowing us to diagnose faithful vs. hallucinatory phases.

\textbf{(2) Micro-level Snapshots:} The bar charts analyze the token selection process for the top candidate tokens at specific steps, by explicitly comparing the model's linguistic preference (its raw logits score) against its visual fidelity.

To power this two-part visualization, our diagnosis hinges on tracking several key metrics (which we introduce here intuitively and define formally in Section~\ref{sec:methodology}):
\begin{itemize}
    \item \textbf{Current Contextual Token Alignment (CCTA):} The CLIP similarity of the image and the latest text window, measuring a new token's immediate impact on visual alignment.
    \item \textbf{Historical Baseline ($\boldsymbol{\bar{B}_t}$):} A smoothed average of recent CCTA scores, reflecting the model's established visual coherence.
    \item \textbf{Isolated Token Alignment (ITA):} A measure of a token's standalone visual relevance, independent of context.
\end{itemize}
Comparing logits against CCTA and ITA allows us to visualize the exact conflict between the model's linguistic bias and the grounded visual evidence.

\textbf{The Diagnosis: Linguistic Priors Hijacking Generation.} Our analysis reveals a clear pattern of behavior. During visually-grounded generation (green phase in Fig~\ref{fig:motivation}), the CCTA trajectory remains high, closely tracking the historical baseline, indicating successful selection of visually coherent tokens. 
However, as the model enters a hallucinatory phase (red phase), a critical divergence occurs. The CCTA score drops sharply, visually confirming the semantic drift. More critically, the surrounding bar charts reveal the direct cause: the model, driven by its powerful linguistic priors, confidently assigns the highest logit score to a visually inappropriate token with low visual alignment. This detrimental selection occurs even when other tokens with far superior visual fidelity are available, which the model disregards, blinded by its linguistic bias.

This investigation pinpoints the root cause of hallucination: a failure of selection. The model’s linguistic priors hijack the decoding process, compelling it to choose a visually unfaithful token even when superior alternatives are present. Standard decoding methods, which primarily rely on the language model's raw logits, are blind to this divergence. They lack the real-time, fine-grained visual consistency check needed to systematically re-evaluate a token's true visual appropriateness against the unfolding narrative. Therefore, we introduce DLC, a framework designed to act as this missing visual referee. Our approach explicitly integrates these real-time visual signals, dynamically assessing each candidate's visual merit against the recent context and adaptively modulating the logits to ensure the generation remains firmly anchored to the visual evidence.

\section{Methodology}
\label{sec:methodology}

\begin{figure}
    \centering
    \includegraphics[width=0.95\linewidth]{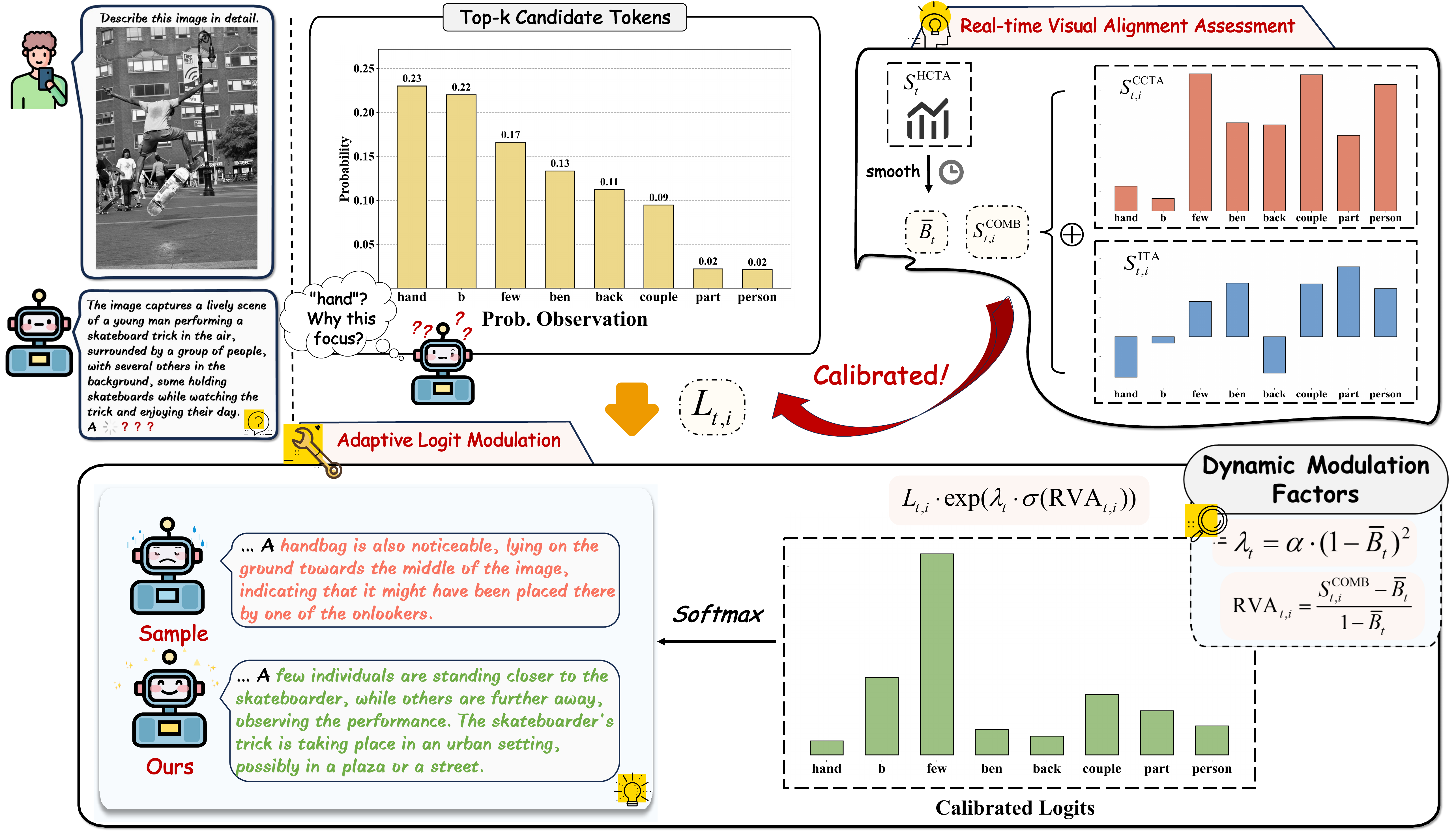}
    \caption{\textbf{Overview of Dynamic Logits Calibration (DLC).} Given an input image and prompt, DLC first performs real-time
    visual alignment assessment on top-k candidate tokens by calculating
    CCTA and ITA scores relative to a $\bar{B}_{t}$. These scores
    inform the adaptive logit modulation step, which computes RVA and $\lambda_{t}$ to adjust the original logits ($L_{t,i}$), favoring visually grounded tokens.}
    \label{fig:overview}
    \vspace{-0.4cm}
\end{figure}

This section details the mechanism of our Dynamic Logits Calibration method. As illustrated in Figure~\ref{fig:overview}, DLC operates at each decoding step by executing two core functions: \textbf{1)} Real-time Visual Alignment Assessment to score the visual fidelity of candidate tokens, and \textbf{2)} Adaptive Logit Modulation to apply corrective pressure to the logits, thereby anchoring the generation to visual evidence.

\subsection{Real-time Visual Alignment Assessment}
\label{ssec:assessment}
At step $t$, DLC first establishes a dynamic benchmark of visual consistency from the recent context, against which it then scores each candidate token.

\textbf{Establishing a Dynamic Visual Baseline.}
To effectively detect semantic deviations, we first establish a dynamic baseline of the model's visual coherence. We maintain a Historical Baseline ($\bar{B}_{t}$), which represents the model's recently established standard of visual fidelity. This baseline is calculated as a smoothed average of the History Contextual Token Alignment (HCTA) scores, computed from a recent window of $N$ tokens. The HCTA score measures the CLIP similarity between the image $v$ and the immediately preceding text window $y_{t-N:t-1}$:
\begin{equation}
\label{eq:hcta}
S_{t}^{HCTA} = \text{CLIP}(v, y_{t-N:t-1})
\end{equation}
The resulting $\bar{B}_{t}$ provides a stable, adaptive reference point against which the visual merit of candidate tokens is judged.

\textbf{Two-Pronged Candidate Assessment.}
For each of the top-$k$ candidate tokens $\{c_1, \dots, c_k\}$, we conduct a two-pronged visual assessment to determine its fidelity. 
\textbf{First}, we perform a contextual check, which addresses the intuitive question: \textit{``Does this token make sense now, given the immediate context?''}
Formally, the CCTA score ($S_{t,i}^{CCTA}$) measures the visual alignment of the sequence resulting from appending $c_i$ to the preceding context:
\begin{equation}
\label{eq:ccta}
S_{t,i}^{CCTA} = \text{CLIP}(v, y_{t-N:t-1} \oplus c_i)
\end{equation}
where $\oplus$ denotes concatenation.
\textbf{Second}, we perform an intrinsic check, which asks: \textit{``Is this token even related to the image at all, in isolation?''} The ITA score ($S_{t,i}^{ITA}$) measures the token's standalone visual relevance:
\begin{equation}
\label{eq:ita}
S_{t,i}^{ITA} = \text{CLIP}(v, c_i)
\end{equation}
This ITA check is a crucial safeguard, especially if the preceding context has already started to drift.


\textbf{Robust Combined Assessment.}
Relying on either check alone is insufficient. The intrinsic check (ITA) acts as a crucial safety brake against flawed context (as discussed in Section~\ref{sec:motivation}), but it is context-blind: it cannot distinguish between multiple visually present objects (e.g., ``dog'' and ``park''), ignoring the narrative logic. 
Conversely, the contextual check (CCTA) provides this essential narrative coherence. It understands that `A dog is in a park' is correct, while `A dog is in a dog' is nonsensical, even if ``dog'' is visually present.

To make a comprehensive judgment, we must integrate these two complementary signals. We employ simple averaging as a principled, parameter-free strategy to balance contextual consistency and intrinsic relevance. This linear combination gives equal voting power to both aspects and ensures that a token is penalized if it fails either check, acting as a robust regularizer. As empirically validated in our ablation studies, removing either the CCTA or ITA component leads to a significant degradation in performance , confirming that this balanced combination is essential.
\begin{equation}
\label{eq:comb}
S_{t,i}^{COMB} = (S_{t,i}^{CCTA} + S_{t,i}^{ITA}) / 2
\end{equation}


\subsection{Adaptive Logit Modulation}
\label{ssec:modulation}

With a robust visual assessment for each candidate, DLC's second function intervenes by adaptively modulating the original LVLM logits to favor visually grounded tokens.

\textbf{Quantifying Relative Visual Advantage (RVA).}
We first quantify the \text{RVA}, which measures the degree to which a candidate token $c_i$ improves or degrades visual alignment relative to the historical baseline:
\begin{equation}
\label{eq:rva}
\text{RVA}_{t,i} = (S_{t,i}^{COMB} - \bar{B}_{t}) / (1 - \bar{B}_{t})
\end{equation}
A positive \text{RVA} indicates the token pulls the generation towards visual truth and vice versa. This metric provides a clear, actionable signal for intervention.

\textbf{Determining the Adaptive Intervention Strength.}
A key design principle of DLC is that intervention should be minimally invasive. Therefore, We introduce $\lambda_t$ as a non-linear control mechanism that adaptively scales the intervention's intensity based on the context's alignment:
\begin{equation}
\label{eq:lambda}
\lambda_t = \alpha \cdot (1 - \bar{B}_{t})^2
\end{equation}
where $\alpha$ is a hyperparameter controlling the maximum strength. A simple linear factor $(1 - \bar{B}_{t})$ would be overly sensitive to minor, noisy fluctuations in the baseline. By squaring the error term $(1 - \bar{B}_{t})$, the intervention strength $\lambda_t$ becomes quadratically small for minor drifts, effectively ignoring insignificant noise. Conversely, as the context's alignment catastrophically drops, the penalty scales assertively. This non-linear control ensures that intervention is applied gently for high-confidence generation but forcefully only when semantic drift is unambiguously detected. As empirically validated in Figure~\ref{fig:ablation_components}, this adaptive factor significantly outperforms a constant intervention strength.

\textbf{Applying the Calibrated Logits.}
Finally, we apply the corrective force, informed by the token's visual merit (RVA) and the overall context alignment ($\lambda_t$), directly onto the original logit $L_{t,i}$. We employ a multiplicative modulation strategy specifically designed to scale the original logit's confidence based on its assessed visual fidelity:
\begin{equation}
\label{eq:final_logit}
L'_{t,i} = L_{t,i} \cdot \exp(\lambda_t \cdot \sigma(\text{RVA}_{t,i}))
\end{equation}
where $\sigma(\cdot)$ is the sigmoid function. 
This multiplicative form allows DLC to dynamically amplify the likelihood of visually advantageous tokens and suppress disadvantageous ones, rather than simply adding an offset. The $\exp(\cdot)$ function ensures this scaling factor remains strictly positive. Furthermore, applying the sigmoid function $\sigma(\cdot)$ to the RVA maps its potentially unbounded value to a smooth, bounded range [0, 1] before scaling by $\lambda_t$. This ensures a stable and controlled application of the visual guidance, preventing overly aggressive adjustments and preserving the nuances of the original language model distribution. This adaptive scaling mechanism effectively allows the visual evidence to win the tug-of-war against misleading linguistic priors, steering the generation back towards visual fidelity.

\section{Experiments}

\begin{table}[t]
    \centering
    \caption{\textbf{CHAIR hallucination evaluation results.} This table evaluates model hallucination using the CHAIR$_{S}$ and CHAIR$_{I}$ metrics, where lower values indicate reduced hallucination and improved performance. \textbf{Bold} and \underline{underlined} values indicate the best and second-best results, respectively.}
    \label{tab:chair_results_full}
    \setlength{\tabcolsep}{8pt} 
    \begin{tabular}{cl cc cc cc}
        \toprule
        \multirow{2}{*}{\begin{tabular}[c]{@{}c@{}}\textbf{Max} \\ \textbf{Token}\end{tabular}} & \multirow{2}{*}{\textbf{Methods}} & \multicolumn{2}{c}{\textbf{LLaVA-1.5}} & \multicolumn{2}{c}{\textbf{InstructBLIP}} & \multicolumn{2}{c}{\textbf{MiniGPT-4}} \\ 
        \cmidrule(lr){3-4} \cmidrule(lr){5-6} \cmidrule(lr){7-8} 
        & & \multicolumn{1}{c}{$C_{S}$ $\downarrow$} & \multicolumn{1}{c}{$C_{I}$ $\downarrow$} & \multicolumn{1}{c}{$C_{S}$ $\downarrow$} & \multicolumn{1}{c}{$C_{I}$ $\downarrow$} & \multicolumn{1}{c}{$C_{S}$ $\downarrow$} & \multicolumn{1}{c}{$C_{I}$ $\downarrow$} \\
        \midrule
        
        \multirow{5}{*}{64}
        & Nucleus & 25.2 & 9.11 & 31.3 & 11.3 & 25.5 & 9.72 \\
        & VCD  & 24.9 & 8.15 & 27.8 & 9.56 & \underline{22.5} & 8.27 \\
        & ICD  & 24.0 & 8.12 & 29.3 & 9.63 & 22.8 & 8.24 \\
        & SID  & \underline{21.9} & \underline{6.84} & \underline{27.7} & \underline{9.14} & 23.7 & \underline{8.21} \\

        \rowcolor{blue!10}
        \cellcolor{white} & DLC(Ours) & \textbf{20.9} & \textbf{6.22} & \textbf{25.4} & \textbf{8.38} & \textbf{22.3} & \textbf{7.84} \\ 
        \midrule
        
        \multirow{5}{*}{512}
        & Nucleus & 52.3 & 16.4 & \underline{56.2} & 17.6 & 35.2 & 12.2 \\
        & VCD  & 56.5 & 17.0 & 62.0 & 18.1 & \underline{31.9} & \underline{10.4} \\
        & ICD  & 52.1 & 15.4  & 65.4 & 19.7 & 33.2 & 10.5 \\
        & SID  & \underline{51.1} & \underline{14.4} & 58.9 & \underline{16.8} & 33.5 & 11.1 \\
        
        \rowcolor{blue!10}
        \cellcolor{white} & DLC(Ours) & \textbf{38.4} & \textbf{10.8} & \textbf{51.8} & \textbf{15.2} & \textbf{31.6} & \textbf{10.0} \\
        \bottomrule
    \end{tabular}
    \vspace{-0.3cm}
\end{table}

\subsection{Experimental Settings}

\textbf{Evaluated LVLMs.} We evaluate the effectiveness of our proposed DLC on three state-of-the-art LVLMs: LLaVA-1.5 \cite{llava}, InstructBLIP \cite{InstructBLIP} and MiniGPT-4 \cite{minigpt4}. All benchmarked models adopt a dual-modality architecture, incorporating pre-trained vision encoders like CLIP \cite{clip} vision encoder and language models (\textit{e.g.}, Vicuna \cite{vicuna}, LLaMA \cite{llama2}) as fundamental components. Notably, all of the models used in our paper are equipped with a 7B LLMs.

\textbf{Benchmarks.} To evaluate our proposed method, we employed a suite of established benchmarks. These include: \textbf{(1) CHAIR} \cite{chair} and \textbf{(2) POPE} \cite{pope}, for quantifying object hallucination in image captions; \textbf{(3) SHR Evaluations} \cite{hadpo}, for detecting more fine-grained hallucination types; \textbf{(4) GPT-4o assisted evaluation} \cite{woodpecker}, for a comprehensive analysis of hallucinations and text quality; \textbf{(5) MME benchmark} \cite{mme}, for assessing the LVLM’s general ability; \textbf{(6) LLaVA-Bench-in-the-Wild} \cite{llava}, for performing targeted case studies on complex tasks. Appendix provides detailed descriptions of the benchmarks.

\textbf{Baselines.} To evaluate our training-free LVLM decoding strategies, we have compared them with standard decoding methods and several state-of-the-art decoding approaches: VCD \cite{vcd}, ICD \cite{icd}, SID \cite{sid}, and OPERA \cite{opera}. For comprehensive comparisons, we implemented VCD, ICD, SID, and DLC using both nucleus sampling (Top-$p$=1). The reported performance of these baselines is based on our re-implementation using their publicly available code.

\subsection{Experimental Results}

\textbf{CHAIR Evaluation.} Following \cite{opera,sid,icd}, we randomly sample 500 images from COCO \cite{mscoco} validation and prompt the evaluated models with ``Please describe this image in detail'' to generate descriptions under two different maximum new-token limits. Moreover, to address CHAIR's \cite{chair} image sensitivity, three distinct image sets were used for robust evaluation. Table \ref{tab:chair_results_full} shows our method generally surpasses others, validating its efficacy. Notably, our method also exhibits superior generation speed over other elaborately designed approaches. Moreover, as our approach mitigates semantic drift, its superiority is particularly evident with the 512-token setting.

\textbf{POPE Evaluation.} To assess the generalizability of our method beyond open-ended descriptive generation, we extended DLC to the VQA domain. We evaluated its performance on the POPE benchmark \cite{pope}, which is designed to probe object existence hallucinations. As presented in Table \ref{tab:pope_results_f1}, DLC achieves state-of-the-art results among training-free methods across all three challenging settings: random, popular, and adversarial. This robust performance demonstrates the effectiveness and versatility of our calibration framework, confirming its applicability to more constrained, reasoning-based tasks.

\begin{table}[t]
    \centering
    \caption{\textbf{Performance on POPE Benchmark.} Results show DLC's effectiveness when extended to VQA tasks. F1-scores are reported for different setups. \textbf{Bold} and \underline{underlined} values indicate the best and second-best results per column, respectively.}
    \label{tab:pope_results_f1}
    \small 
    \setlength{\tabcolsep}{3pt} 
    
    \begin{tabular}{@{}l ccc ccc ccc@{}}
        \toprule
        \multirow{2}{*}{\textbf{Method}} & \multicolumn{3}{c}{\textbf{LLaVA-1.5}} & \multicolumn{3}{c}{\textbf{InstructBLIP}} & \multicolumn{3}{c}{\textbf{MiniGPT-4}} \\
        \cmidrule(lr){2-4} \cmidrule(lr){5-7} \cmidrule(lr){8-10}
        & \textit{Rand.} & \textit{Pop.} & \textit{Adv.} & \textit{Rand.} & \textit{Pop.} & \textit{Adv.} & \textit{Rand.} & \textit{Pop.} & \textit{Adv.} \\
        \midrule
        Nucleus       & 84.23          & 81.85          & 77.83          & 83.43          & 78.78          & 77.02          & 63.25          & 60.79          & 59.99 \\
        ICD           & 85.48          & 82.80          & 78.39          & 83.68          & 77.66          & 76.15          & 62.43          & 60.39          & 59.47 \\
        VCD           & 83.20          & 81.13          & 77.66          & 83.69          & 78.68          & 76.99          & 64.40          & 61.90          & 61.36 \\
        SID           & 87.10          & 84.11          & 80.43          & 84.43          & 78.98          & 78.07          & 64.25          & 61.46          & 60.65 \\
        \midrule
        DLC+CLIP-336      & \underline{88.81} & \textbf{85.99} & \textbf{80.86} & \underline{89.01} & 82.46          & \textbf{81.15} & 75.19          & 70.73          & 68.46 \\
        DLC+SigLIP    & 88.14          & 85.23          & 79.68          & 88.81          & \textbf{83.69} & 80.82          & \textbf{78.16} & \underline{72.72} & \underline{70.38} \\
        DLC+FG-CLIP   & \textbf{88.88} & \underline{85.84} & \underline{80.68} & \textbf{89.32} & \underline{83.42} & \underline{81.10} & \underline{77.97} & \textbf{72.73} & \textbf{70.45} \\
        \bottomrule
    \end{tabular}
    \vspace{-0.4cm}
\end{table}

\begin{figure}[t]
    \centering
    \includegraphics[width=0.99\linewidth]{
        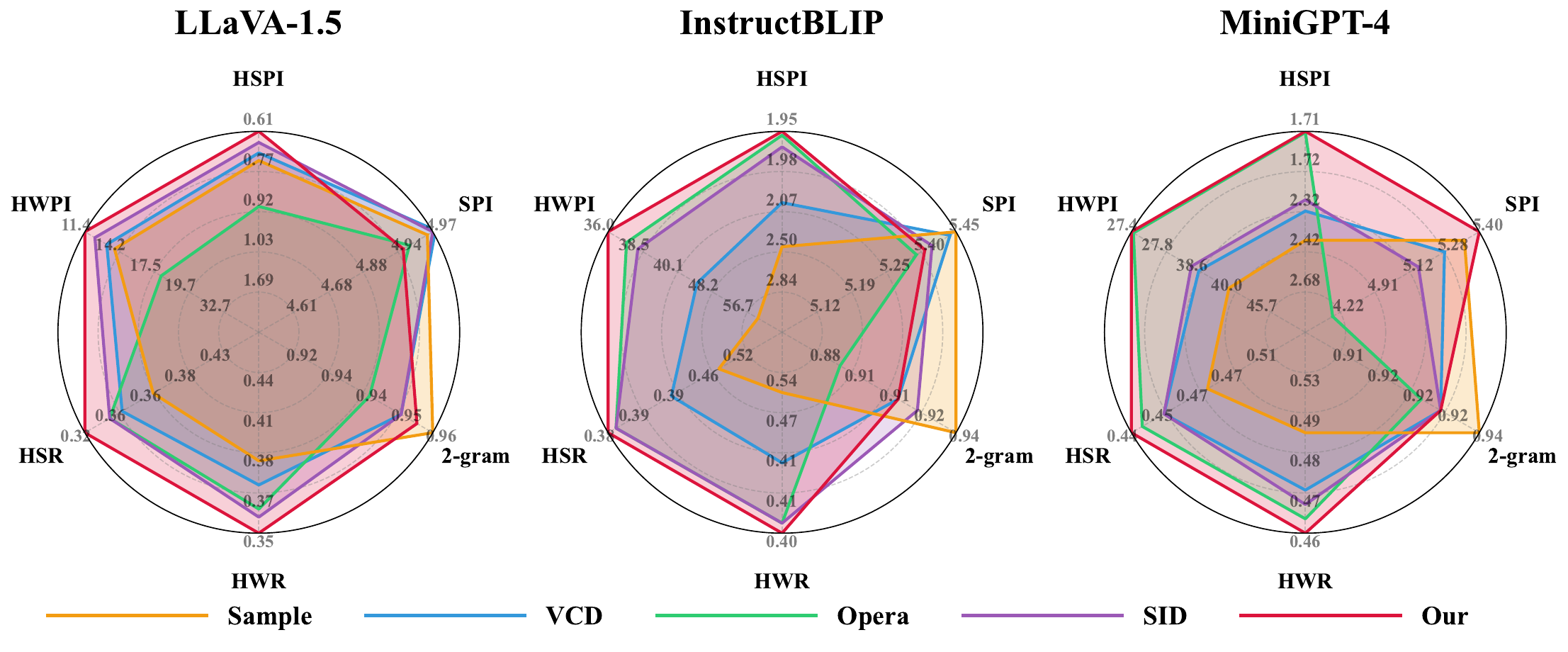
    }
    \caption{\textbf{SHR evaluation results.} Six aspects are analyzed, including the number of sentences per image (SPI), fluency and local smoothness (2-gram), the number of hallucinated sentences per
    image (HSPI), the number of hallucinated words per image (HWPI), the ratio
    of hallucinated sentences (HSR), and the ratio of hallucinated words (HWR). Larger radar indicates better performance.}
    \label{fig:gpt4}
    \vspace{-0.5cm}
\end{figure}

\textbf{SHR Evaluation.} Following \cite{opera,sid,hadpo}, we evaluate on 200 images from the VG dataset using the prompt ``Please describe this image in detail'' with a maximum of 512 new tokens generated.
As illustrated in Figure \ref{fig:gpt4}, DLC achieves state-of-the-art performance on hallucination-related metrics at both the sentence and word levels, indicating its effectiveness in mitigating hallucinations driven by strong semantic priors.
To characterize generation quality beyond faithfulness, we additionally report a n-gram fluency score ($n{=}2$) as a proxy for local smoothness and the number of generated sentences per image (SPI) to compare the detailedness of generated texts.
Consistent with observations in OPERA \cite{opera}, we find that for certain models, DLC may reduce the length of generated descriptions, which is mainly attributable to suppressing superfluous hallucinatory content rather than degrading the core visual description.

\textbf{GPT-4o Assisted Evaluation.} To comprehensively analyze hallucination and text quality across diverse data distributions, we employ GPT-4o as an advanced evaluation proxy to score Correctness ($C$) and Detailedness ($D$). Our evaluation spans two distinct benchmarks: standard object descriptions on COCO \cite{mscoco} (500 randomly selected images) and the more challenging, out-of-distribution scenarios in LLaVA-Bench-in-the-Wild \cite{llava}.The quantitative results, consolidated in Table \ref{tab:gpt4v_all}, benchmark DLC against Nucleus Sampling and established decoding strategies including VCD \cite{vcd}, OPERA \cite{opera}, and SID \cite{sid}. The unified view reveals that DLC achieves remarkable consistency: it outperforms baselines across both benchmarks and all three model architectures. Notably, DLC demonstrates significant gains in correctness over conventional sampling while maintaining comparable descriptive detail. This robustness across datasets suggests that DLC is not merely overfitting to specific object distributions but offers a generalized improvement in grounding. Furthermore, we provide an in-depth qualitative analysis of case studies in Appendix. These cases visually demonstrate how the DLC approach markedly reduces hallucinatory content and enhances descriptive detail under identical image and prompt conditions.

\begin{table}[t]
    \centering
    \caption{\textbf{Hallucination Evaluation with GPT-4o.} We report Correctness and Detailedness across two benchmarks. Grouping by model allows for direct comparison of performance consistency across COCO and LLaVA-Bench-in-the-Wild (L-Bench).}
    \label{tab:gpt4v_all}
    \resizebox{\linewidth}{!}{
        \setlength{\tabcolsep}{2.5pt}
        \begin{tabular}{@{}l cccc cccc cccc@{}}
            \toprule
            \multirow{3}{*}{\textbf{Settings}} 
            & \multicolumn{4}{c}{\textbf{LLaVA-1.5}} 
            & \multicolumn{4}{c}{\textbf{InstructBLIP}} 
            & \multicolumn{4}{c}{\textbf{MiniGPT-4}} \\
            \cmidrule(lr){2-5} \cmidrule(lr){6-9} \cmidrule(lr){10-13}
            
            & \multicolumn{2}{c}{COCO} & \multicolumn{2}{c}{L-Bench} 
            & \multicolumn{2}{c}{COCO} & \multicolumn{2}{c}{L-Bench} 
            & \multicolumn{2}{c}{COCO} & \multicolumn{2}{c}{L-Bench} \\
            \cmidrule(lr){2-3} \cmidrule(lr){4-5} 
            \cmidrule(lr){6-7} \cmidrule(lr){8-9} 
            \cmidrule(lr){10-11} \cmidrule(lr){12-13}
            
            & $C \uparrow$ & $D \uparrow$ & $C \uparrow$ & $D \uparrow$ 
            & $C \uparrow$ & $D \uparrow$ & $C \uparrow$ & $D \uparrow$ 
            & $C \uparrow$ & $D \uparrow$ & $C \uparrow$ & $D \uparrow$ \\
            \midrule
            
            Nucleus 
            & 5.34 & 6.18 & 4.71 & 5.79  
            & 4.76 & 5.61 & 4.21 & 5.00  
            & 5.15 & 5.73 & 5.17 & 5.33 \\ 
            
            Our 
            & \textbf{7.73} & \textbf{6.87} & \textbf{8.42} & \textbf{7.04} 
            & \textbf{7.38} & \textbf{6.75} & \textbf{7.25} & \textbf{6.42} 
            & \textbf{6.64} & \textbf{6.56} & \textbf{6.00} & \textbf{5.75} \\
            
            \midrule
            
            VCD 
            & 5.55 & 6.38 & 5.25 & 5.96 
            & 5.32 & 6.09 & 4.88 & 5.54 
            & 5.55 & 5.97 & 5.00 & 5.08 \\
            
            Our 
            & \textbf{7.45} & \textbf{6.67} & \textbf{7.46} & \textbf{6.58} 
            & \textbf{6.94} & \textbf{6.54} & \textbf{6.58} & \textbf{6.17} 
            & \textbf{6.39} & \textbf{6.44} & \textbf{5.67} & \textbf{5.46} \\
            
            \midrule
            
            OPERA 
            & 6.16 & 6.59 & 5.75 & 6.75 
            & 6.02 & 6.17 & 5.88 & 5.62 
            & 5.98 & 5.93 & 5.75 & \textbf{6.08} \\
            
            Our 
            & \textbf{7.10} & \textbf{6.67} & \textbf{7.54} & \textbf{6.88} 
            & \textbf{6.51} & \textbf{6.55} & \textbf{5.96} & \textbf{6.21} 
            & \textbf{6.16} & \textbf{6.45} & \textbf{5.79} & 5.62 \\
            
            \midrule
            
            SID 
            & 5.69 & 6.54 & 5.58 & \textbf{6.79} 
            & 5.34 & 6.07 & 4.42 & 5.25 
            & 5.55 & 6.05 & 5.38 & 5.42 \\
            
            Our 
            & \textbf{7.17} & \textbf{6.55} & \textbf{6.71} & 6.08 
            & \textbf{6.89} & \textbf{6.62} & \textbf{6.38} & \textbf{6.29} 
            & \textbf{6.54} & \textbf{6.45} & \textbf{5.88} & \textbf{5.79} \\
            
            \bottomrule
        \end{tabular}
    }
    \vspace{-0.2cm}
\end{table}

\begin{figure}[htbp]
    \centering
    \includegraphics[width=0.99\linewidth]{
        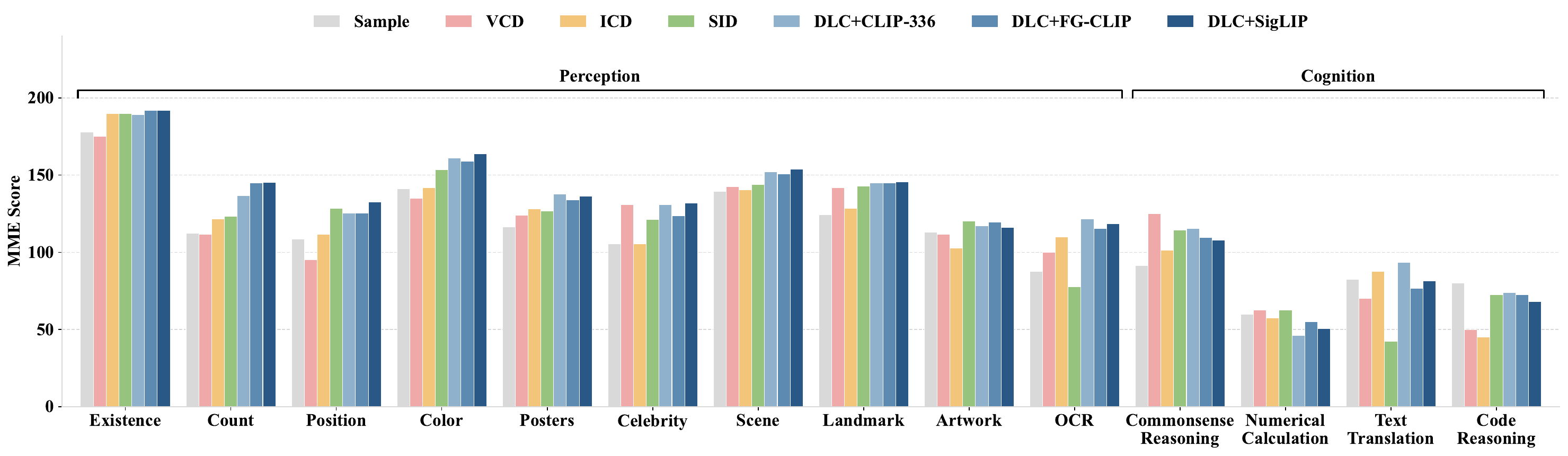
    }
    \caption{\textbf{Comprehensive breakdown of MME benchmark performance on LLaVA-1.5.} We compare DLC variants against standard decoding baselines across 14 subtasks spanning perception and cognition suites.}
    \label{fig:mme_result}
    \vspace{-0.5cm}
\end{figure}

\textbf{Broad Capabilities on MME.} To verify that our hallucination mitigation strategy does not compromise general multimodal capabilities, we conducted a comprehensive evaluation on the MME benchmark \cite{mme}. The results, illustrated in Figure \ref{fig:mme_result}, reveal a nuanced trade-off.
In Perception tasks, DLC achieves consistent gains, corroborating its effectiveness in correcting visual grounding errors.
In Cognition tasks, performance varies by type. For visually-grounded reasoning, DLC improves over the baseline, as clearer perception aids understanding. However, for abstract symbolic tasks such as Code Reasoning and Numerical Calculation, we observe a minor performance regression compared to the unconstrained base model. This suggests that enforcing strict visual alignment may impose a slight constraint on pure symbolic reasoning. Notably, however, DLC avoids the catastrophic degradation seen in contrastive baselines, maintaining a much more balanced capability profile.

\subsection{Ablation and Analysis}

\begin{table}[t]
\centering
\caption{\textbf{Efficiency benchmarking and sensitivity analysis.} We report end-to-end inference time (s/sample) and GPU memory usage on LLaVA-1.5 (NVIDIA RTX 3090). The upper section compares DLC with standard baselines; the lower section analyzes the trade-off between candidate pool size $k$, latency, and faithfulness.}
\label{tab:efficiency_tradeoff}
\small
\setlength{\tabcolsep}{6pt}
\begin{tabular}{lcccc}
\toprule
\textbf{Method} & \textbf{Time} $\downarrow$ & \textbf{Memory (MB)} $\downarrow$ & \textbf{CHAIR$_s$} $\downarrow$ & \textbf{CHAIR$_i$} $\downarrow$ \\
\midrule
Vanilla & 3.64 & 15062 & 56.6 & 18.2 \\
VCD & 5.22 & 15062 & 59.4 & 18.9 \\
OPERA & 7.43 & 20854 & 53.7 & 15.0 \\
\midrule
\textbf{DLC} ($k{=}10$)  & 4.84 & 17874 & 41.2 & 11.2 \\
\textbf{DLC} ($k{=}30$)  & 5.11 & 17874 & 38.2 & 10.6 \\
\textbf{DLC} ($k{=}50$)  & 5.35 & 17868 & \textbf{35.8} & \underline{10.5} \\
\textbf{DLC} ($k{=}100$) & 6.25 & 17862 & \underline{36.6} & \textbf{10.3} \\
\textbf{DLC} ($k{=}200$) & 9.09 & 18118 & 39.8 & 11.5 \\
\bottomrule
\end{tabular}
 \vspace{-0.3cm}
\end{table}

\begin{figure}[htbp]
    \centering
    \includegraphics[width=0.99\linewidth]{
        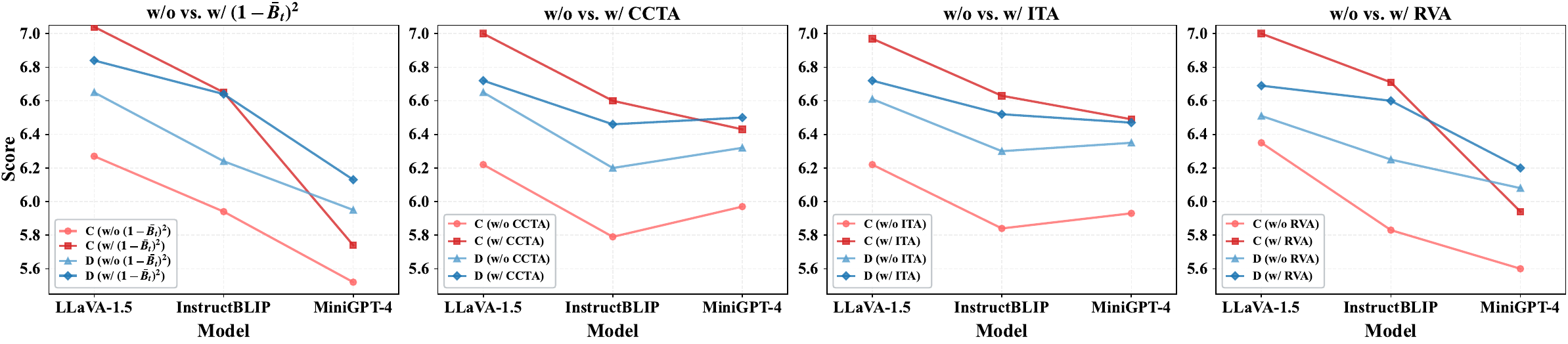
    }
    \caption{\textbf{Ablation studies on DLC core components.} GPT-4o evaluates Correctness (C) and Detailedness (D) on the COCO dataset across three LVLMs.}
    \label{fig:ablation_components}
    \vspace{-0.5cm}
\end{figure}

In this section, we present a systematic analysis of DLC. 
Specifically, we (i) ablate key components to verify their individual contributions, 
(ii) study the effect of the candidate pool size $k$ and its quality--cost trade-off in runtime and memory, and 
(iii) evaluate robustness to different visual backbones and scalability to larger LVLMs. 
We include further results in the Appendix, such as hyperparameter sensitivity and evaluations under different decoding strategies.

\textbf{Effectiveness of Core Components.}
To isolate the contribution of each core component, we performed ablation studies evaluating model correctness and detailedness on the COCO dataset. The results are visualized in Figure \ref{fig:ablation_components}. We validated the efficacy of: (1) the adaptive factor in the dynamic guidance strength $\lambda_{t}$, (2) the Current Contextual Token Alignment (CCTA), (3) the Isolated Token Alignment (ITA), and (4) the Relative Visual Advantage (RVA). As the figure clearly demonstrates, removing or replacing any of these components leads to a noticeable degradation in performance across all three tested LVLMs. These findings confirm that each component is integral to DLC's effectiveness.

\textbf{Efficiency and Sensitivity Analysis.}
Table \ref{tab:efficiency_tradeoff} presents a dual analysis: comparing DLC's efficiency against established baselines and dissecting the impact of the candidate pool size $k$.
DLC demonstrates a superior efficiency-quality profile. Notably, at a moderate setting, DLC is faster than VCD while drastically reducing hallucination.
Meanwhile, the ablation reveals a decoupled scaling behavior. Inference time increases linearly with $k$ due to the computational cost of scoring more candidates. However, GPU memory usage remains remarkably stable regardless of $k$. This confirms that the memory overhead is static rather than dynamic, making DLC memory-friendly even with larger candidate pools.
Furthermore, performance does not scale monotonically. While increasing $k$ from 10 to 50 significantly improves grounding by capturing more relevant candidates, pushing $k$ to extremes yields diminishing returns and even degrades performance. This degradation likely stems from long-tail noise—introducing too many low-probability tokens that confuse the selection mechanism. Consequently, we recommend $k \in [30, 50]$ as the optimal range, offering state-of-the-art faithfulness with inference speeds comparable to or faster than contrastive baselines.

\begin{table}[t]
    \centering
    \begin{minipage}[t]{0.54\linewidth}
        \centering
        \caption{\textbf{CHAIR Results on 13B Models.} Comparison of nucleus sampling baseline and DLC on larger model architectures.}
        \label{tab:chair_13b_results}

        \resizebox{\linewidth}{!}{
            \setlength{\tabcolsep}{3.5pt} 
            \begin{tabular}{@{}l c cc cc@{}}
                \toprule
                \multirow{2}{*}{Methods} & \multirow{2}{*}{\begin{tabular}[c]{@{}c@{}}Max \\ Token\end{tabular}} & \multicolumn{2}{c}{LLaVA-1.5} & \multicolumn{2}{c}{InstructBLIP} \\
                \cmidrule(lr){3-4} \cmidrule(lr){5-6}
                & & $C_{S}$ $\downarrow$ & $C_{I}$ $\downarrow$ & $C_{S}$ $\downarrow$ & $C_{I}$ $\downarrow$ \\
                \midrule
                Nucleus & 64 & 23.2 & 8.16 & 27.8 & 10.2 \\
                DLC & 64 & \textbf{18.2} & \textbf{5.50} & \textbf{24.2} & \textbf{7.75} \\
                \midrule
                Nucleus & 512 & 53.2 & 15.1 & 64.0 & 19.6 \\
                DLC & 512 & \textbf{33.0} & \textbf{8.80} & \textbf{55.6} & \textbf{15.4} \\
                \bottomrule
            \end{tabular}
        }
    \end{minipage}
    \hfill 
    \begin{minipage}[t]{0.45\linewidth}
        \centering
        \caption{\textbf{GPT-4o Evaluation on 13B Models.} Assessment of correctness and detailedness.}
        \label{tab:gpt4o_13b_results}
        \resizebox{\linewidth}{!}{
            \setlength{\tabcolsep}{3.5pt}
            \begin{tabular}{@{}l cc cc@{}}
                \toprule
                \multirow{2}{*}{Methods} & \multicolumn{2}{c}{LLaVA-1.5} & \multicolumn{2}{c}{InstructBLIP} \\ 
                \cmidrule(lr){2-3} \cmidrule(lr){4-5}
                & $C \uparrow$ & $D \uparrow$ & $C \uparrow$ & $D \uparrow$ \\
                \midrule
                Nucleus  & 5.25 & 6.05 & 4.26 & 5.06 \\
                DLC & \textbf{7.75} & \textbf{6.77} & \textbf{7.00} & \textbf{6.36} \\
                \midrule
                VCD & 5.93 & 6.44 & 5.27 & 5.80 \\
                DLC & \textbf{7.32} & \textbf{6.63} & \textbf{6.86} & \textbf{6.09} \\
                \bottomrule
            \end{tabular}
        }
    \end{minipage}
    \vspace{-0.2cm}
\end{table}

\begin{table}[h]
    \centering
    \caption{\textbf{Robustness analysis of DLC with different CLIP backbones.} Performance is evaluated on the CHAIR benchmark across three LVLMs.}
        \vspace{-0.2cm}
    \label{tab:clip_robustness}
    \begin{tabular}{lcccccc@{}}
        \toprule
        \multirow{2}{*}{Settings} & \multicolumn{2}{c}{LLaVA-1.5} & \multicolumn{2}{c}{InstructBLIP} & \multicolumn{2}{c}{MiniGPT-4} \\
        \cmidrule(lr){2-3} \cmidrule(lr){4-5} \cmidrule(lr){6-7}
        & $C_{S}$ $\downarrow$ & $C_{I}$ $\downarrow$ & $C_{S}$ $\downarrow$ & $C_{I}$ $\downarrow$ & $C_{S}$ $\downarrow$ & $C_{I}$ $\downarrow$ \\
        \midrule
        CLIP-336 & \textbf{36.6}    & \textbf{10.2}     & \textbf{49.4}     & \textbf{12.4} &\textbf{30.8} &10.2     \\
        SigLIP  & 39.4    & 10.9     & 51.0     & 14.3 & 32.0 & 11.2    \\
        FG-CLIP & 37.8  & 10.7 & 51.2   & 13.9 &31.6 &\textbf{9.87} \\
        \bottomrule
    \end{tabular}
    \vspace{-0.3cm}
\end{table}

\textbf{Robustness Analysis Across Different CLIP Models.} To analyze the robustness of our DLC method and its dependence on the underlying CLIP model, we conducted comprehensive experiments using different CLIP backbone. Specifically, we tested three models: the CLIP-336 (clip-vit-large-patch14-336) \cite{clip}, SigLIP (siglip-so400m-patch14-384) \cite{siglip}, and FG-CLIP (fg-clip-large) \cite{fgclip}. The results in Table \ref{tab:pope_results_f1} Table \ref{tab:clip_robustness} and Figure \ref{fig:mme_result} demonstrate that DLC consistently improves performance across all combinations, validating its robustness.

\textbf{Scalability to Larger Model Architectures.}
We further evaluate DLC on 13B variants of LLaVA-1.5 and InstructBLIP (Tables \ref{tab:chair_13b_results} and \ref{tab:gpt4o_13b_results}). 
DLC consistently outperforms baselines across both metrics: it significantly reduces object hallucination in CHAIR evaluation while simultaneously improving holistic correctness and detailedness scores in GPT-4o assessments. These findings confirm that DLC generalizes effectively to stronger backbones without specific tuning.
\section{Conclusion}
This paper tackles \emph{semantic drift}---a dynamic failure mode where linguistic priors progressively hijack LVLM generation and detach it from visual evidence.
We argue that hallucination is best understood as a token-level trajectory failure: as drift emerges, the model can still place grounded candidates among the top options, yet selects a linguistically plausible but visually unfaithful token.
To address this root cause, we introduced DLC, a training-free decoding framework that injects explicit visual grounding signals into token selection in real time.
DLC acts as a visual referee by combining intrinsic relevance and contextual coherence checks and comparing them to an adaptive historical baseline, thereby computing a relative visual advantage used to dynamically calibrate logits.
Across multiple LVLMs and diverse benchmarks, DLC consistently reduces hallucinations while maintaining response quality and broad multimodal capabilities.
Further analyses on different vision backbones and the candidate pool size reveal a favorable quality--cost trade-off with predictable overhead.
We hope our findings on semantic drift and our decoding-time grounding mechanism will inspire future work on more reliable and trustworthy multimodal generation.


%
%
\bibliographystyle{splncs04}
\bibliography{main}

\clearpage
\setcounter{page}{1}

\section{Supplementary Material}
This appendix provides supplementary materials to support the main findings of our paper. 
We begin with additional background on LVLMs in Section~\ref{app:background}. 
Section~\ref{app:benchmarks} provides further details on the benchmarks used in our experiments. 
We then present the detailed pseudo-code for DLC in Section~\ref{app:pseudo_code} and elaborate on our specific implementation details in Section~\ref{app:impl_details}.

The remainder of the appendix is dedicated to extended experimental results and analyses.
Section~\ref{app:hyperparams} presents a sensitivity analysis for our key hyperparameters, $\alpha$ and $N$.
Section~\ref{app:decoding_strategies} shows DLC's compatibility with various other decoding strategies.
Finally, Section~\ref{app:prompts} provides the exact prompts used for our GPT-4 and GPT-4o evaluations, and Section~\ref{app:llavabench} presents additional qualitative case studies from the LLaVA-Bench \cite{llava} dataset.

\subsection{More backgrounds on LVLMs}\label{app:background} 
The development of LVLMs has evolved from BERT-based \cite{bert,vilbert,lxmert} cross-modal alignment to LLM-powered \cite{flamingo,shikra,InstructBLIP,Blip-2,llava,minigpt4} architectures. Building upon foundational text-image alignment techniques like CLIP \cite{clip} and BLIP \cite{blip}, contemporary LVLMs integrate visual encoders with LLM decoders through distinct optimization strategies. For instance, LLaVA \cite{llava} employs lightweight adapters combined with instruction tuning, while InstructBLIP \cite{InstructBLIP} and MiniGPT-4 \cite{minigpt4} utilize Q-Former modules for visual token compression. Despite these innovations, significant challenges persist, particularly severe hallucination issues arising from misaligned multimodal representations. To systematically analyze mitigation strategies, we have examined three representative architectures, including LLaVA, InstructBLIP, and MiniGPT-4, whose results demonstrate that various technical approaches commonly share critical vulnerabilities to the hallucination issue.

\subsection{Further Details on Benchmarks }\label{app:benchmarks}
To comprehensively evaluate our DLC method, we utilized several benchmarks specifically designed to assess hallucination in vision-language models. This section provides additional details about these benchmarks, including dataset characteristics, evaluation protocols, and metrics interpretation.

    (1) \textbf{CHAIR} \cite{chair} quantifies object hallucination
    in image captions by measuring the ratio of objects mentioned in captions but
    absent from ground-truth labels. It includes two variants: CHAIR\textsubscript{I} ($C_I$),
    calculating hallucinated objects versus all generated objects, and CHAIR\textsubscript{S} ($C_S$),
    assessing sentences with hallucinations versus total sentences. These two metrics are defined as follows:
\begin{equation}
    \begin{split}
        C_I &= \frac{|\{\text{hallucinated objects}\}|}{|\{\text{all mentioned objects}\}|}, \\
        C_S &= \frac{|\{\text{captions with hallucinated objects}\}|}{|\{\text{all captions}\}|}
    \end{split}
\end{equation}
    Lower scores indicate better hallucination suppression. 

    (2) \textbf{POPE} (Polling-based Object Probing Evaluation) \cite{pope} is a benchmark designed to directly and efficiently assess object-level hallucinations in LVLMs. It operates by querying the model with a series of simple polling-based (Yes/No) questions about the presence or absence of specific objects within an image. Based on the MS-COCO \cite{mscoco} dataset, POPE generates both positive samples and, more importantly, challenging negative samples to rigorously test a model's susceptibility to hallucination.
    Random: An object is randomly selected from all COCO categories to serve as a baseline test of factual grounding.
    Popular: An object with a high frequency of occurrence across the dataset is selected. This challenges models that rely heavily on statistical priors, tempting them to affirm the existence of common objects regardless of visual evidence.
    Adversarial: An object semantically, functionally, or spatially related to objects present in the image is chosen. For example, if the image contains a tennis racket, the model might be asked about a tennis ball. This probes for subtle, context-driven hallucinations where a model fabricates objects based on strong co-occurrence patterns.
    By evaluating the model's accuracy, precision, recall, and F1-score across these question types, POPE provides a straightforward yet fine-grained metric for quantifying a model's tendency to hallucinate.

    (3) \textbf{SHR Evaluations} \cite{hadpo} function as a crucial complement to conventional assessment methodologies. While traditional metrics such as CHAIR effectively quantify hallucinations at the object-existence level, they demonstrate inadequacy in detecting more fine-grained hallucination types, including positional, relational, and attribute-based inaccuracies. To address this limitation, GPT-4 assisted evaluation employs detailed object-level descriptions from the Visual Genome (VG) \cite{vgdataset} dataset as ground-truth reference, leveraging GPT-4's sophisticated capabilities for precise hallucination detection and classification. The complete prompt template is provided in the Appendix \ref{app:prompts}.

    (4) \textbf{GPT-4o assisted evaluations}\footnote{We utilize GPT-4o for assisted evaluation, as it offers enhanced capabilities over previous vision-capable models like GPT-4V, which is also being phased out by OpenAI.} \cite{woodpecker} further conduct a comprehensive analysis of hallucinations and text quality in open-ended generation tasks. Harnessing GPT-4o's sophisticated capabilities, Yin \etal~\cite{woodpecker} established a numerical
    assessment protocol utilizing a 0-10 scale, which evaluates two essential dimensions: Accuracy-the extent to which responses align with image content without fabrication, and Detailedness-the degree of informational richness present in the responses.

    (5) \textbf{MME Benchmark} \cite{mme} provides a comprehensive assessment of LVLM capabilities by evaluating both perceptual and cognitive skills. The benchmark is structured into a total of 14 subtasks, with ten designed to test perception and four to assess cognition. All evaluations are performed using a yes/no question format. MME features specific sub-benchmarks for hallucination discrimination, including tasks that probe existence, count, position, and color. These tasks are intended to examine both object-level and attribute-level hallucinations. Task scores are used as the metric to report performance on this benchmark.
    
    (6) \textbf{LLaVA-Bench-in-the-Wild} \cite{llava} is designed to evaluate LVLMs' proficiency in addressing complex tasks and their adaptability to novel domains. We perform targeted case studies on this dataset to demonstrate the efficacy of our proposed DLC.

\subsection{Pseudo-code of Our Method}\label{app:pseudo_code}

In this section, we provide a detailed pseudo-code of our Dynamic Logits Calibration (DLC) method, as shown in Algorithm~\ref{alg:dlc}.

\begin{algorithm}[h]
\caption{Dynamic Logits Calibration (DLC)}
\label{alg:dlc}
\begin{algorithmic}[1]
\Require Input image $v$, text prompt $x$, VLM $M_{\theta}$, CLIP model
\Require Window size $N$, modulation strength $\alpha$
\Ensure Generated text sequence $y$
\State $y \gets x$ \Comment{Initialize with prompt}
\State $\bar{B} \gets 0$ \Comment{Initialize historical baseline}
\State $H \gets \text{empty queue}$ \Comment{History of HCTA scores}
\For{$t = |x|+1$ \textbf{to} $|x|+\text{max\_new\_tokens}$}
    \State $L_t \gets M_{\theta}(y_{<t}, v)$ \Comment{Get original logits}
    
    \If{$t > |x|+3$} \Comment{Start DLC after warm-up}
        \State $y_{\text{ctx}} \gets y_{t-N:t-1}$ \Comment{Get preceding context}
        \State $S_t^{\text{HCTA}} \gets \text{CLIP}(v, y_{\text{ctx}})$ \Comment{Calc. HCTA score}
        
        \State Add $S_t^{\text{HCTA}}$ to $H$; maintain size $N$
        
        \State $\bar{B}_t \gets \avg(H)$ \Comment{Update historical baseline}
        
        \State $C \gets \TopK(L_t)$ \Comment{Get top-k candidate tokens}
        
        \For{each candidate token $c_i \in C$}
            \State $S_{t,i}^{\text{CCTA}} \gets \text{CLIP}(v, y_{\text{ctx}} \oplus c_i)$
            \State $S_{t,i}^{\text{ITA}} \gets \text{CLIP}(v, c_i)$
            \State $S_{t,i}^{\text{COMB}} \gets (S_{t,i}^{\text{CCTA}} + S_{t,i}^{\text{ITA}}) / 2$
            
            \State $\text{RVA}_{t,i} \gets (S_{t,i}^{\text{COMB}} - \bar{B}_t) / (1 - \bar{B}_t)$
            \State $\lambda_t \gets \alpha \cdot (1 - \bar{B}_t)^2$ \Comment{Dynamic strength}
            \State $L'_{t,i} \gets L_{t,i} \cdot \exp(\lambda_t \cdot \sigma(\text{RVA}_{t,i}))$ \Comment{Adjust logit}
        \EndFor
        
        \State $L_t \gets L'_t$ \Comment{Update logits with adjusted ones}
    \EndIf
    
    \State $y_t \gets \text{sample}(L_t)$
    \State $y \gets y \oplus y_t$
    
    \If{$y_t$ is end token} \textbf{break} \EndIf
\EndFor
\State \Return $y$
\end{algorithmic}
\end{algorithm}

\subsection{Implementation Details}\label{app:impl_details}
In this section, we provide comprehensive implementation details of our Dynamic Logits Calibration (DLC) method, including model configuration, parameter settings, and computational workflows. These details are essential for reproducing our results and understanding the practical aspects of DLC.

Unless otherwise specified, we compute all visual alignment scores
using the SigLIP model \cite{siglip}. The historical baseline $\bar{B}_{t}$ is
calculated over a sliding window of $N=8$ past context scores, following a 3-step
warm-up period. For the top-k candidates ($k=50$) at each step, the combined score $S_{t,i}^{\text{COMB}}$ is computed. The adaptive modulation employs
a maximum intervention strength $\alpha=3$, and the standard sigmoid function $\sigma
(\cdot)$ for logit adjustment.

\textbf{CLIP Query Efficiency.}
To ensure high throughput, the $k=50$ CLIP queries required at each decoding step are implemented for high efficiency. This efficiency stems from two key factors: \textbf{(1) Batch Processing:} All 50 candidate queries are batched into a single, parallel forward pass through the CLIP text encoder, which is highly optimized on modern GPUs. \textbf{(2) Low Computational Load:} The text input for each query is extremely short. The computational cost of encoding these short snippets is negligible compared to the cost of a full forward pass through the multi-billion parameter LVLM. As a result, this process has a minimal impact on inference speed. This single, lightweight CLIP pass is a core design choice for efficiency.

\subsection{Sensitivity Analysis of Key Hyperparameters: $\alpha$ and $N$}\label{app:hyperparams}
The effectiveness of our DLC method hinges on two key hyperparameters: the modulation strength $\alpha$ and the context window size $N$. This section presents a sensitivity analysis of these parameters, evaluating their impact on hallucination mitigation across three distinct LVLMs. Performance is measured using the CHAIR metric \cite{chair}, with a maximum of 512 new tokens generated for all ablation studies.

\begin{table}[htbp]
    \centering
    \caption{Impact of Modulation Strengths $\alpha$ on CHAIR Metrics for Different LVLMs and Decoding Strategies}
    \label{tab:chair_decoding_vs_alpha_comparison}
    \setlength{\tabcolsep}{5pt}
    \begin{tabular}{@{}l c cc cc cc@{}} 
    \toprule
    \multirow{2}{*}{Method} & \multirow{2}{*}{\textbf{$\alpha$}} & \multicolumn{2}{c}{LLaVA-1.5} & \multicolumn{2}{c}{InstructBLIP} & \multicolumn{2}{c}{MiniGPT-4} \\
    \cmidrule(lr){3-4} \cmidrule(lr){5-6} \cmidrule(lr){7-8} &                      & $C_{S} \downarrow$            & $C_{I} \downarrow$               & $C_{S} \downarrow$          & $C_{I} \downarrow$ & $C_{S} \downarrow$ & $C_{I} \downarrow$ \\
    \midrule 
    \multirow{5}{*}{Greedy}      
                                & 1 & \textbf{33.8} & \textbf{9.85} & \textbf{47.9} & \textbf{13.2} & \textbf{32.6} & 10.3 \\
                                & 2 & 34.1 & 10.0 & 48.1 & 13.5 & 33.7 & 10.1 \\
                                & 3 & 34.8 & 10.1 & 49.6 & 13.5 & 33.0 & \textbf{9.89} \\
                                & 4 & 35.3 & 9.87 & 50.5 & 14.2 & 33.5 & 10.0 \\
                                & 5 & 36.8 & 10.5 & 51.0 & 14.4 & 33.9 & 10.6 \\
    \midrule
    \multirow{5}{*}{Nucleus} 
                                & 1 & 41.8 & 12.7 & 55.4 & 15.4 & 34.6 & 11.1 \\
                                & 2 & 40.3 & 11.4 & 53.2 & 14.8 & 34.0 & \textbf{10.6} \\
                                & 3 & 39.4 & 10.9 & 51.0 & \textbf{14.3} & \textbf{32.0} & 11.2 \\
                                & 4 & \textbf{37.5} & \textbf{10.8} & 50.8 & 14.7 & 33.6 & 10.7 \\
                                & 5 & \textbf{37.5} & 10.9 & \textbf{50.6} & 14.5 & 33.6 & 10.9 \\

    \bottomrule
    \end{tabular}
\end{table}

\begin{table}[h]
    \centering
    \caption{Impact of Window Size $N$ on CHAIR Metrics for Different LVLMs and Decoding Strategies}
    \label{tab:chair_metrics_window_size_beautified}
    \setlength{\tabcolsep}{5pt}
    \begin{tabular}{@{}llcccccc@{}}
        \toprule
        \multirow{2}{*}{Method} & \multirow{2}{*}{$N$} & \multicolumn{2}{c}{LLaVA-1.5} & \multicolumn{2}{c}{InstructBLIP} & \multicolumn{2}{c}{MiniGPT4} \\
        \cmidrule(lr){3-4} \cmidrule(lr){5-6} \cmidrule(lr){7-8}
         & & $C_{S} \downarrow$ & $C_{I} \downarrow$ & $C_{S} \downarrow$ & $C_{I} \downarrow$ & $C_{S} \downarrow$ & $C_{I} \downarrow$ \\
        \midrule
        
        \multirow{4}{*}{Greedy} 
        & 4  & \textbf{34.4} & \textbf{9.45} & 52.2 & 14.6 & 35.4 & 11.1 \\
        & 8  & 35.1 & 9.72 & 49.0 & \textbf{12.8} & 33.4 & 10.3 \\
        & 12 & \textbf{34.4} & 9.51 & 48.0 & 13.5 & 33.2 & 9.85 \\
        & 16 & 37.0 & 9.84 & \textbf{47.0} & 13.2 & \textbf{31.0} & \textbf{9.69} \\
        
        \midrule
        
        \multirow{4}{*}{Nucleus} 
        & 4  & 38.6 & 11.1 & 51.4 & \textbf{14.3} & \textbf{31.0} & 10.2 \\
        & 8  & 39.4 & \textbf{10.9} & 51.0 & \textbf{14.3} & 32.0 & 11.2 \\
        & 12 & 38.6 & 11.5 & 54.8 & 14.8 & 32.4 & \textbf{10.1} \\
        & 16 & \textbf{37.6} & 11.4 & \textbf{50.4} & 15.0 & 34.0 & 12.6 \\
        
        \bottomrule
    \end{tabular}
\end{table}

\textbf{Impact of Modulation Strength ($\alpha$):}
We first examine the modulation strength, $\alpha$. Our findings, detailed in Table~\ref{tab:chair_decoding_vs_alpha_comparison}, reveal that the optimal $\alpha$ setting is highly dependent on both the LVLM and the decoding strategy.

With Greedy Search, lower $\alpha$ values generally yield better performance. Specifically, $\alpha=1$ delivers the strongest results for LLaVA-1.5 and InstructBLIP. For MiniGPT-4, performance is competitive at both low ($\alpha=1$) and moderate ($\alpha=3$) strengths.

With Nucleus Sampling, the trend reverses, with moderate-to-high $\alpha$ values (from 3 to 5) consistently improving CHAIR scores. Optimal performance is observed around $\alpha=4$ for LLaVA-1.5, $\alpha=3$ for MiniGPT-4, and $\alpha=5$ for InstructBLIP.

These results underscore that while a global value of $\alpha=3$ is used for our main experiments, tailoring this parameter to the specific model and decoding method can unlock further performance gains.

\textbf{Impact of Context Window Size ($N$):}
Next, we analyze the influence of the context window size, $N$. As shown in Table~\ref{tab:chair_metrics_window_size_beautified}, the optimal choice of $N$ also varies significantly across different configurations.

With Greedy Decoding, InstructBLIP and MiniGPT-4 generally benefit from larger window sizes, achieving their best scores at $N=16$. In contrast, the effect on LLaVA-1.5 is less monotonic, with smaller to moderate window sizes ($N=4$ or $N=12$) proving more advantageous.

With Nucleus Sampling, the response is more varied. MiniGPT-4 performs best with a small window ($N=4$), showing a decline in performance as $N$ increases. LLaVA-1.5's performance fluctuates without a clear trend, while InstructBLIP shows no significant improvement from changes in $N$.

While we adopt a uniform $N=8$ for our main experiments, this analysis highlights that model- and strategy-specific tuning of the window size is crucial for maximizing performance.

In summary, both modulation strength $\alpha$ and window size $N$ are influential hyperparameters for the DLC method. The ablation studies presented here clearly demonstrate that while the default values used in this paper provide a strong general baseline, optimal results are achieved by tuning these parameters for each LVLM and decoding strategy combination.

\subsection{DLC Performance with Various Decoding Strategies}\label{app:decoding_strategies}
\begin{figure*}[ht]
        \centering
        \includegraphics[width=0.95\linewidth]{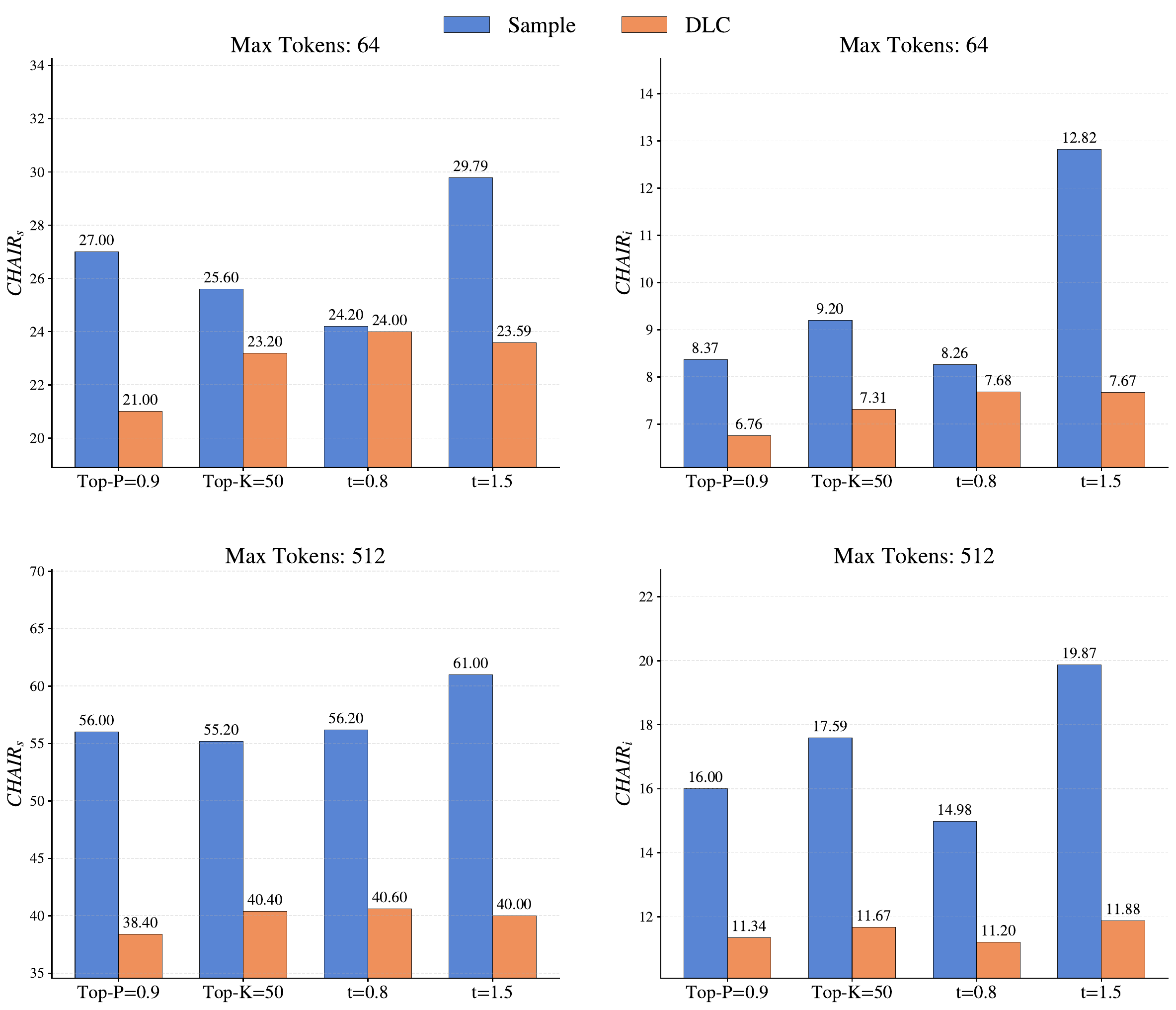}
        \caption{\textbf{Results of different decoding strategies.}}
        \label{fig:decoding_comparison}
    \end{figure*}

Our proposed DLC demonstrates compatibility with various mainstream decoding strategies. To comprehensively evaluate its effectiveness, in addition to its performance with common strategies such as nucleus sampling and greedy sampling, we conducted further experiments on the LLaVA-1.5 \cite{llava} model using the MSCOCO dataset \cite{mscoco} and the CHAIR benchmark \cite{chair}. These experiments explored other decoding methods and parameters, specifically: Top-p sampling (p=0.9), Top-k sampling (k=50), and Top-k sampling (k=50) with varying temperatures (t=1.5 and t=0.8). The results, presented in Figure~\ref{fig:decoding_comparison}, consistently indicate that regardless of the sampling method employed, DLC effectively reduces hallucinations and enhances overall model performance. This substantiates the robustness and effectiveness of our proposed DLC across diverse decoding strategies.

\subsection{Prompts for GPT-4 and GPT-4o Evaluation}\label{app:prompts}
\begin{figure*}[htbp]
        \centering
        \includegraphics[width=0.75\linewidth]{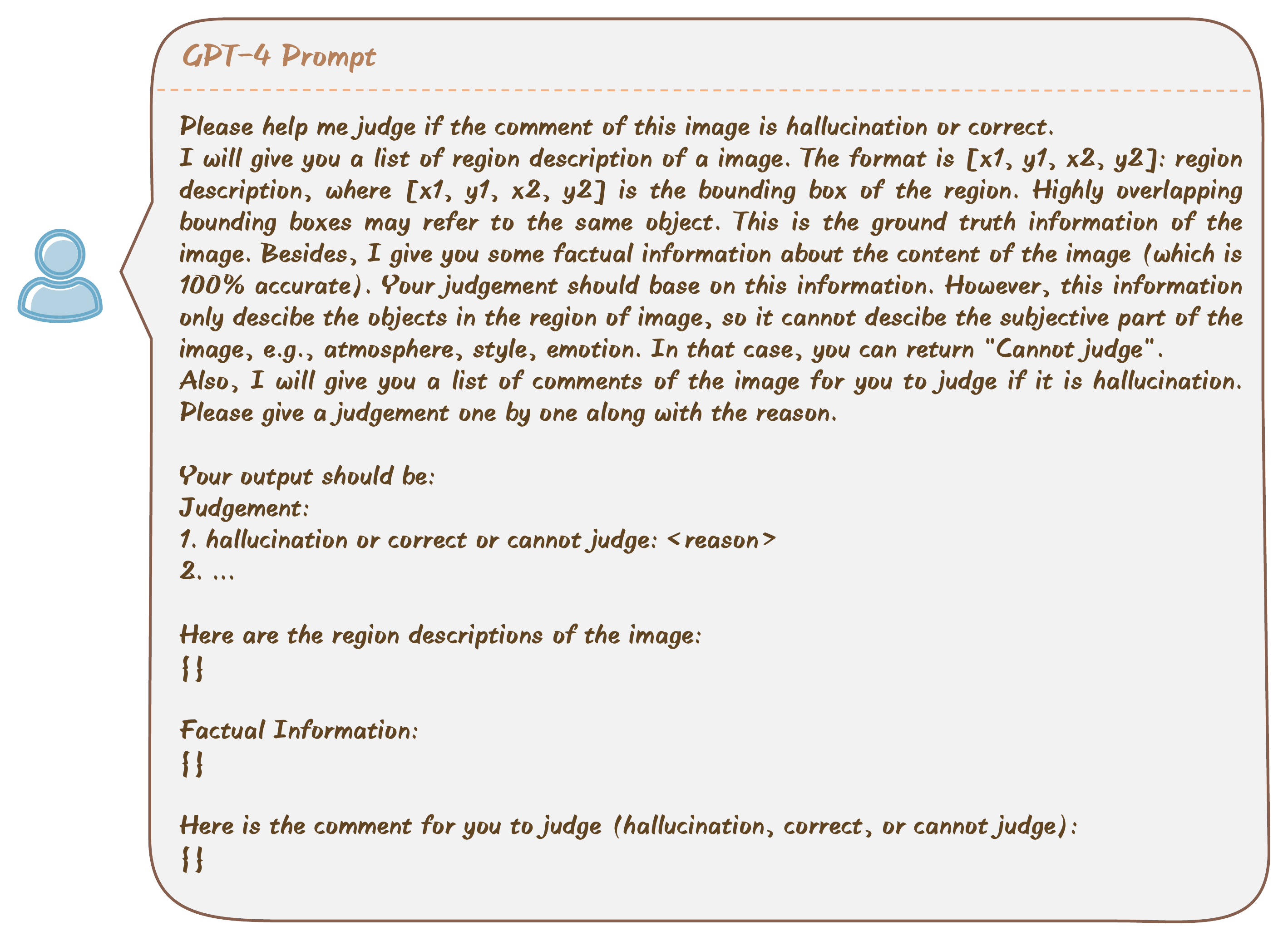}
        \caption{\textbf{Prompts of GPT-4 for evaluations.}}
        \label{fig:gpt4_prompt}
    \end{figure*}

\begin{figure*}[htbp]
        \centering
        \includegraphics[width=0.75\linewidth]{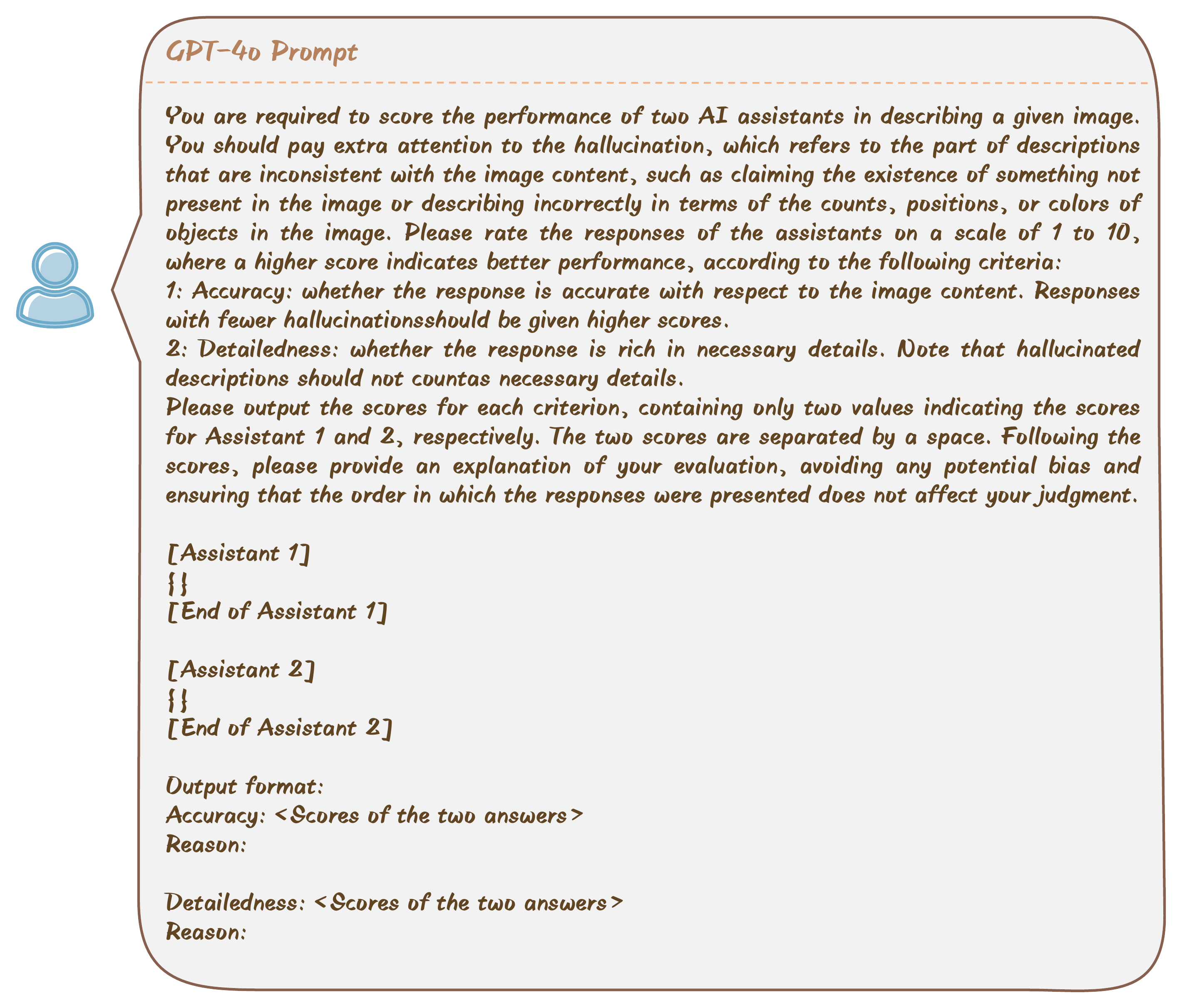}
        \caption{\textbf{Prompts of GPT-4o for evaluations.}}
        \label{fig:gpt4o_prompt}
    \end{figure*}

In this section, we provide the prompts for GPT-4 and GPT-4o evaluation, as shown in Figure~\ref{fig:gpt4_prompt} and Figure~\ref{fig:gpt4o_prompt}.

\subsection{Extended Comparative Results on LLaVA-Bench}\label{app:llavabench}
In this section, we present additional comparative results of our DLC approach against other methods on the LLaVA-Bench \cite{llava}. These extended evaluations provide a more comprehensive understanding of our method's performance across various question types and scenarios.
    \begin{figure*}[htbp]
        \centering
        \includegraphics[width=0.99\linewidth]{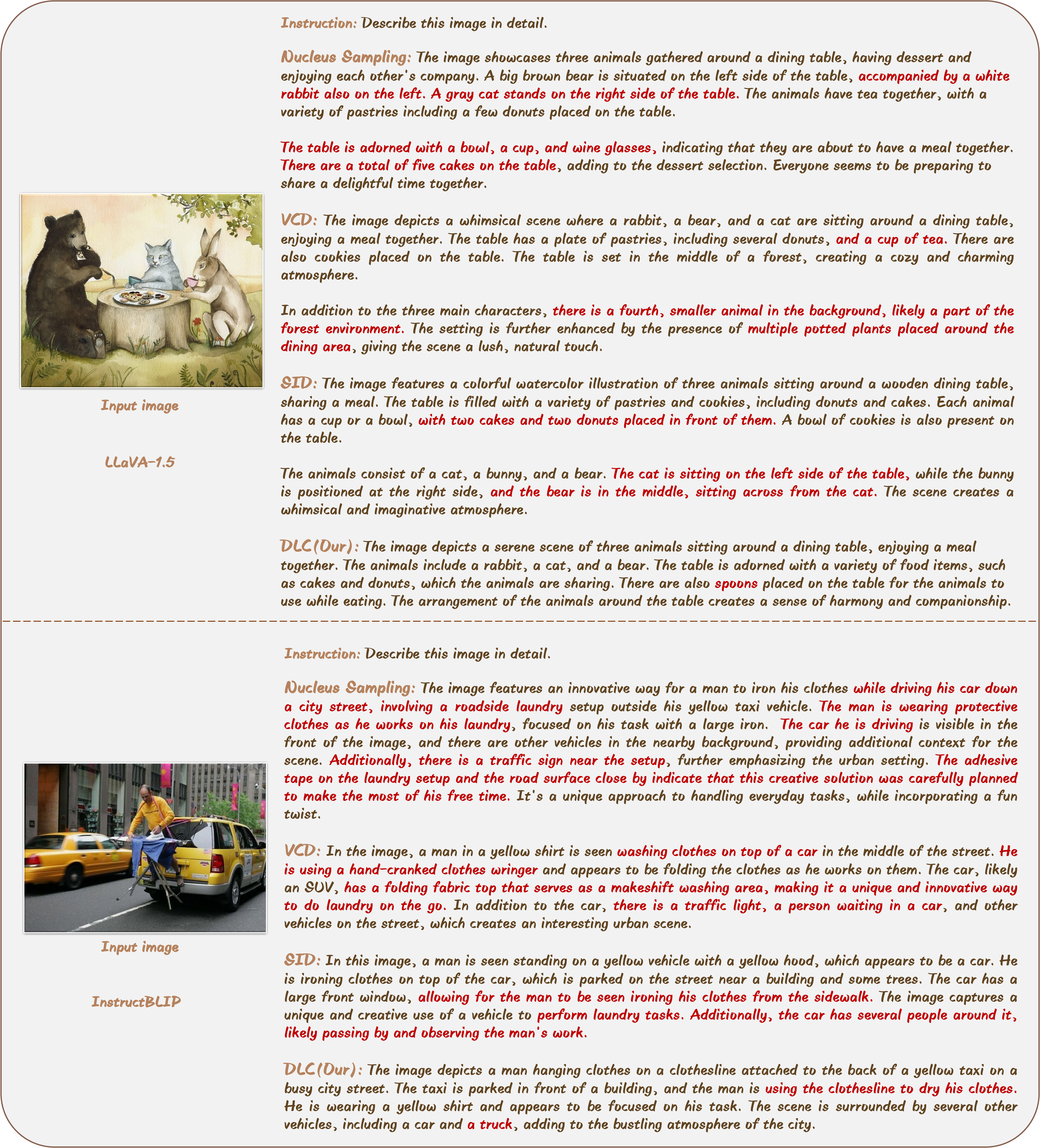}
        \caption{\textbf{Case Study:} LLaVA-1.5 and InstructBLIP on LLaVA-Bench-in-the-Wild (Hallucinations Marked Red).}
        \label{fig:case_combine}
    \end{figure*}
    \begin{figure*}[htbp]
        \centering
        \includegraphics[width=0.99\linewidth]{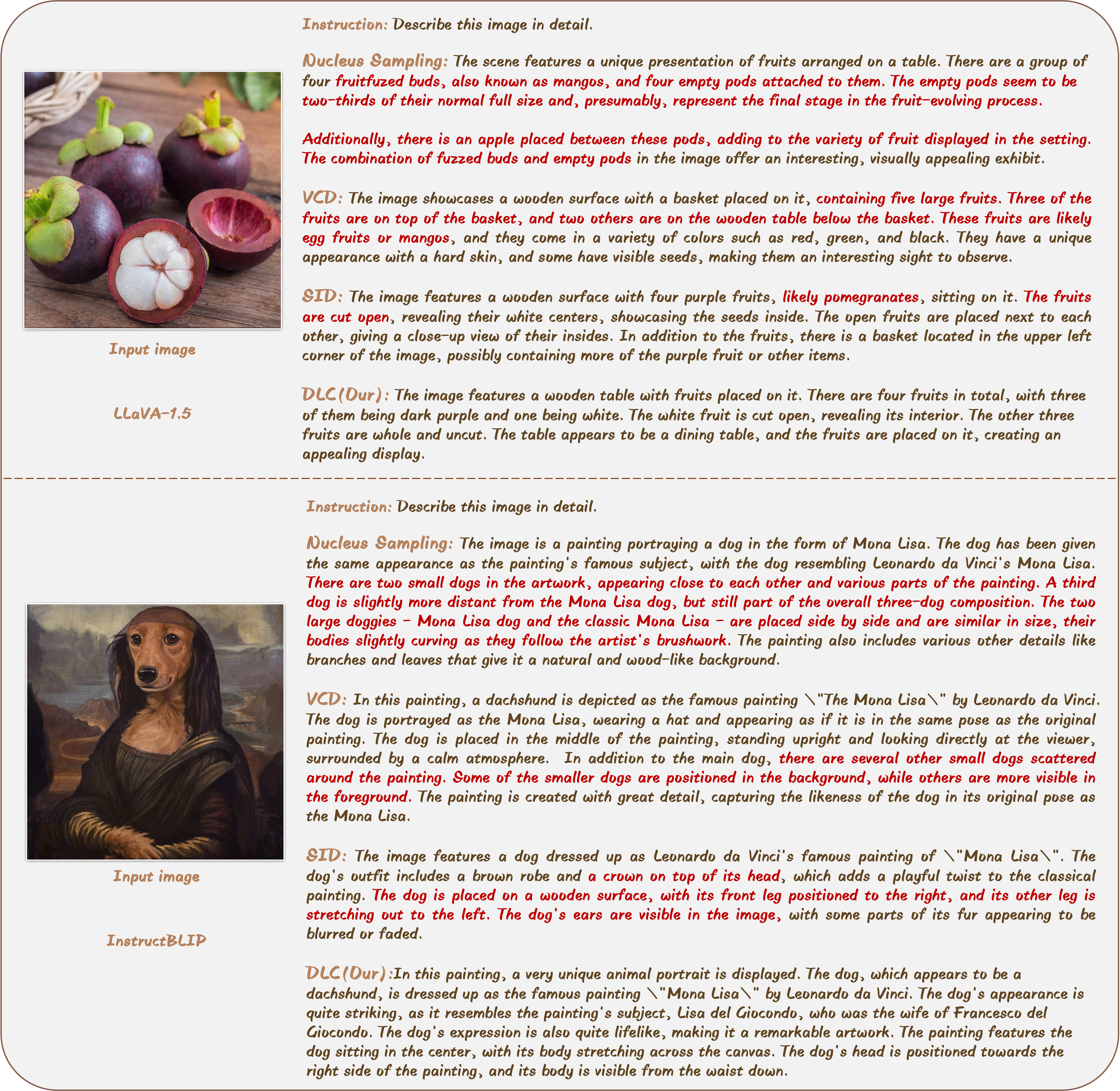}
        \caption{\textbf{Case Study:} LLaVA-1.5 and InstructBLIP on LLaVA-Bench-in-the-Wild (Hallucinations Marked Red).}
        \label{fig:case_combine}
    \end{figure*}

\end{document}